\documentclass[12pt]{elsarticle}
\title{The NN-Stacking: Feature weighted linear stacking through neural networks}

\author[1,2]{Victor Coscrato\corref{cor1}%
}\ead{vcoscrato@gmail.com}
\author[1,2,3]{Marco Henrique de Almeida In\'acio}
\ead[url]{marcoinacio.com}
\author[1]{Rafael Izbicki}
\ead{rafaelizbicki@gmail.com}

\cortext[cor1]{Corresponding author}
\address[1]{Federal University of S\~ao Carlos, Rodovia Washington Lu\'is, Km 235, S\~ao Carlos - SP, Brazil}
\address[2]{ICMC/USP, Av. Trabalhador S\~ao Carlense, 400 - Centro, S\~ao Carlos - SP, Brazil}
\address[3]{TMIT/BME, 1111 Budapest, Műegyetem rkp. 3, Hungary}

\usepackage[T1]{fontenc}
\usepackage{nameref}
\usepackage[pdfencoding=unicode, psdextra, colorlinks=true, urlcolor=blue, citecolor=blue, linkcolor=blue]{hyperref}
\usepackage{amsfonts, amsmath, amssymb, amsthm, thmtools}
\usepackage{mathrsfs}
\usepackage{cleveref}
\usepackage{indentfirst}
\usepackage[bottom=2cm, top=2cm, left=3cm, right=3cm]{geometry}
\usepackage{url}
\usepackage{diagbox}
\usepackage{multirow}
\usepackage{wrapfig}
\usepackage{tikz}
\usepackage{adjustbox}

\numberwithin{equation}{section}

\declaretheorem[name=Theorem, refname={Theorem, Theorems}, Refname={Theorem, Theorems}, parent=section]{theorem}

\declaretheorem[name=Example, refname={Example, Examples}, Refname={Example, Examples}, sibling=theorem, style=definition]{example}

\crefname{section}{section}{sections}
\Crefname{section}{Section}{Sections}
\crefname{table}{table}{tables}
\Crefname{table}{Table}{Tables}

\usepackage{color, enumitem, fancyhdr, float, fourier, layout, multirow, setspace, verbatim}
\setlist[enumerate]{leftmargin=*}
\usepackage[colorinlistoftodos]{todonotes}
\usepackage{listings}
\usepackage{caption, subcaption}

\usepackage{algorithm}
\usepackage{algpseudocode}
\renewcommand{\algorithmicrequire}{\textbf{\small Input:}}
\renewcommand{\algorithmicensure}{\textbf{\small Output:}}

\usepackage[numbers]{natbib}
\usepackage{enumitem}
\usepackage{graphicx}
\graphicspath{{figures/}{../figures/}}

\def\x{{\vec{x}}}

\def\X{{\vec{X}}}

\def\y{{\vec{y}}}

\def\E{{\textbf{E}}}

\def\t0{{\theta_0}}

\def\1{{\boldmath{1}}}

\usepackage{MnSymbol}

\renewcommand{\vec}[1]{\mathbf{#1}}

\usepackage{pgfplots}
\pgfplotsset{compat=1.14}

\begin{document}

\begin{abstract}
Stacking methods improve the prediction performance of regression models. A simple way to stack base regressions estimators is by combining them linearly, as done by \citet{breiman1996stacked}. Even though this approach is useful from an interpretative perspective, it often does not lead to high predictive power. 
We propose the NN-Stacking method (NNS), which generalizes Breiman's method by allowing the linear parameters to vary with input features. This improvement enables NNS to take advantage of the fact that 
distinct base models often perform better at different regions of the feature space.
Our method uses neural networks to estimate the stacking coefficients. We show that while our approach keeps the interpretative features of Breiman's method at a local level, it leads to better predictive power, especially in datasets with large sample sizes.

\end{abstract}

\begin{keyword}
meta-learning \sep neural networks \sep model stacking \sep model selection
\end{keyword}

\maketitle

\section{Introduction}

The  standard procedures for model selection in prediction problems is  cross-validation and data splitting. However, such an approach is known to be sub-optimal \citep{dvzeroski2004combining, dietterich2000ensemble, sill2009feature}. The  reason  is that one might achieve more accurate predictions by \emph{combining} different regression estimators rather then by \emph{selecting} the best one. Stacking methods \citep{zhou2012ensemble} are a way of overcoming such a drawback from standard model selection.

A well known stacking method was introduced by \citet{breiman1996stacked}. This approach consists in taking a linear combination of base regression estimators. That is, the stacked regression has the shape $\sum_{i = 1}^k \theta_i g_i(\x)$, where $g_i$'s are the individual regression estimators (such as random forests, linear regression or support vector regression), $\theta_i$ are weights that are estimated from data and $\x$ represents the features.

Even though this linear stacking method leads to combined estimators that are  easy to interpret, 
it may be sub-optimal in cases where models have different \emph{local accuracy}, i.e., situations where the performance of these estimators vary over the feature space. Example \ref{ex:local} illustrates this situation.

\begin{example}
\label{ex:local}
Consider predicting $Y$ based on a single feature, $x$, using the data in Figure \ref{fig:local}. We fit two least squares estimators: $g_1(\x) = \theta_{01} + \theta_{11} x$  and $g_2(\x) = \theta_{02} + \theta_{12} x^2$. None of the models is uniformly better; for example, the linear fit has better performance when $x \leq 5$, but the quadratic fit yields better performance for $x \in (-2.5, 2.5)$. One may take this into account when creating the stacked estimator
by assigning different weights for each regression according to $x$: while one can assign a larger weight to the linear fit on the regime $x \leq 5$, a lower weight should be assigned to it if $x \in (-2.5, 2.5)$.

\begin{figure}[!htb]
\centering
\includegraphics[scale=0.7]{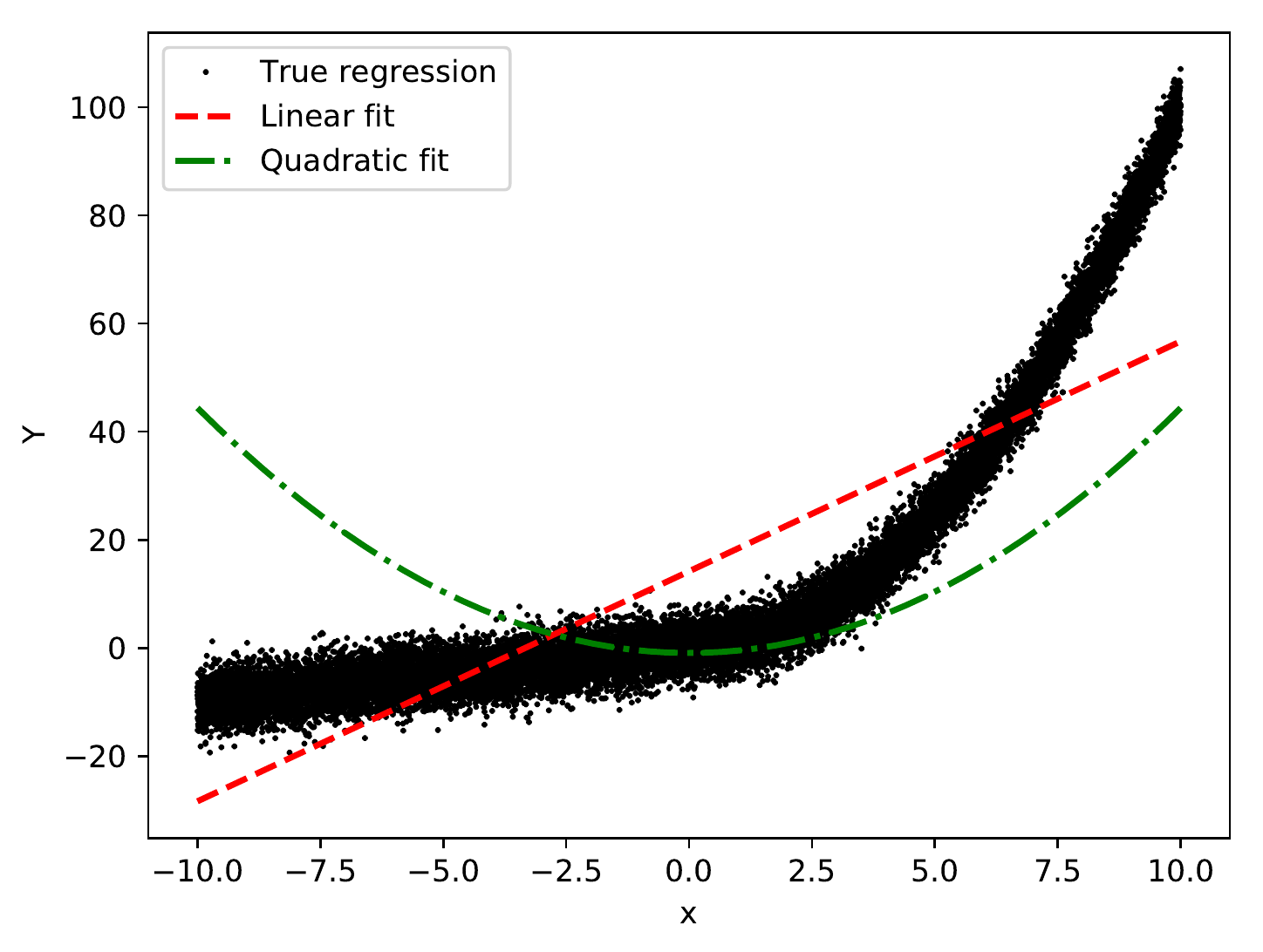}
\caption{Regressions comparison. While for some regions of $x$ the linear fit outperforms the quadratic fit,  in other regions the opposite happens.}
\label{fig:local}
\end{figure}
\end{example}

It is well known that different
regression methods may perform better on different regions of the feature
space. For instance, because local estimators do not suffer from boundary effects, they achieve good performance closer to the edges of the feature space \citep{fan1992variable}. Random forests, on the other hand, implicitly perform feature selection, and thus may have better performance in regions where some features are not relevant \citep{breiman2001random}.

In this work we improve Breiman's approach so that it can take local accuracy into account. That is, we develop a meta-learner that is able to learn which models have higher importance on each region of the feature space. We achieve this goal by allowing each parameter $\theta_i$ to vary  as a function of the features $\x$. In this way, the meta-learner can adapt to each region of the feature space, which  yields higher predictive power. Our approach
keeps the local interpretability of the linear stacking model. 
 
The remaining of the work is organized as follows.  Section \ref{sec:breiman}  introduces the notation
used in the paper, as well as our method. Section \ref{sec:solution} shows  details on its implementation.  Section \ref{sec:app} shows applications of our method to a variety of datasets to evaluate its performance. Section \ref{sec:conclusion} concludes the paper.

\section{Notation and Motivation}
\label{sec:breiman}
The stacking method proposed by \citet{breiman1996stacked} 
 is a  linear combination of $k$ regression functions
for a  label $Y \in \mathbb{R}$.
More precisely,
let $g_\x = (g_1(\x), g_2(\x), \ldots, g_k(\x))'$ be a vector of regression estimators, that is, $g_i(\x)$ is an estimate of $\E[Y|\x], \forall \, i = 1,2,\ldots,k$. The linear  stacked regression is defined as
\begin{align}
\label{eq:breiman}
G_\theta(\x) := \sum_{i = 1}^k \theta_i g_i(\x) = \theta'g_\x
\end{align}
where $\theta = (\theta_1,\theta_2,\ldots,\theta_k)'$ are meta-parameters. One way to estimate the  meta-parameters
using data $(\x_1,y_1),\ldots,(\x_n,y_n)$ is through the least squares method, computed using a leave-one-out setup:
\begin{align}
\label{eq:least_squares}
\arg \min_\theta \sum_{i = 1}^n (y_i - G^{(-i)}_\theta(\x_i))^2 = \arg \min_\theta \sum_{i = 1}^n (y_i - \theta' g_{\x_i}^{(-i)})^2,
\end{align}
where 
$g_j^{(-i)}(\x_i)$ is the prediction for $\x_i$ made by the $j$-th regression fitted without the $i$-th instance.
Note that it is important to use this hold-out approach because if the base regression functions  $g_1(\x), g_2(\x), \ldots, g_k(\x)$ are constructed using the same data as $\theta_1,\ldots,\theta_k$, this can cause $G_\theta(\x)$ to over-fit the training data. 

In order for the stacked estimator to be easier to interpret, 
\citet{breiman1996stacked} also 
requires $\theta_i$'s to be weights, that is $\theta_i \geq 0 \, \forall \, i = 1,2,\ldots,k$ and that $\sum_{i=1}^k \theta_i = 1$.

Even though Breiman's solution
works on a  variety of settings, it does not take into account that each regression method may perform
better in distinct regions of the feature space. In order to overcome this limitation,
we propose  the \emph{Neural Network Stacking} (NNS) which
generalizes Breiman's approach by allowing
$\theta$ on Equation \ref{eq:breiman} to vary  with $\x$. 
That is, our meta-learner has the shape
\begin{align}
\label{eq:ours}
G_\theta(\x) := \sum_{i = 1}^k \theta_i(\x) g_i(\x) = \theta_x'g_\x,
\end{align}
where $\theta_x := (\theta_1(\x),\theta_2(\x),\ldots,\theta_k(\x))'$. 
In other words, the NNS is a \emph{local linear} meta-learner.
Example \ref{ex:local2} 
shows that NNS can substantially
decrease the prediction error of  Breiman's approach.

\begin{example}
\label{ex:local2}
We fit both Breiman's linear meta-learner and our NNS local linear meta-learner
to the models fitted in Example \ref{ex:local}. Figure \ref{fig:local2} shows that 
Breiman's meta-learner is not able to fit the true regression satisfactorily
because
both estimators have poor performance on specific regions of the data. On the other hand,  feature-varying weights yield a better fit.

\begin{figure}[!ht]
    \centering
    \includegraphics[scale=0.75]{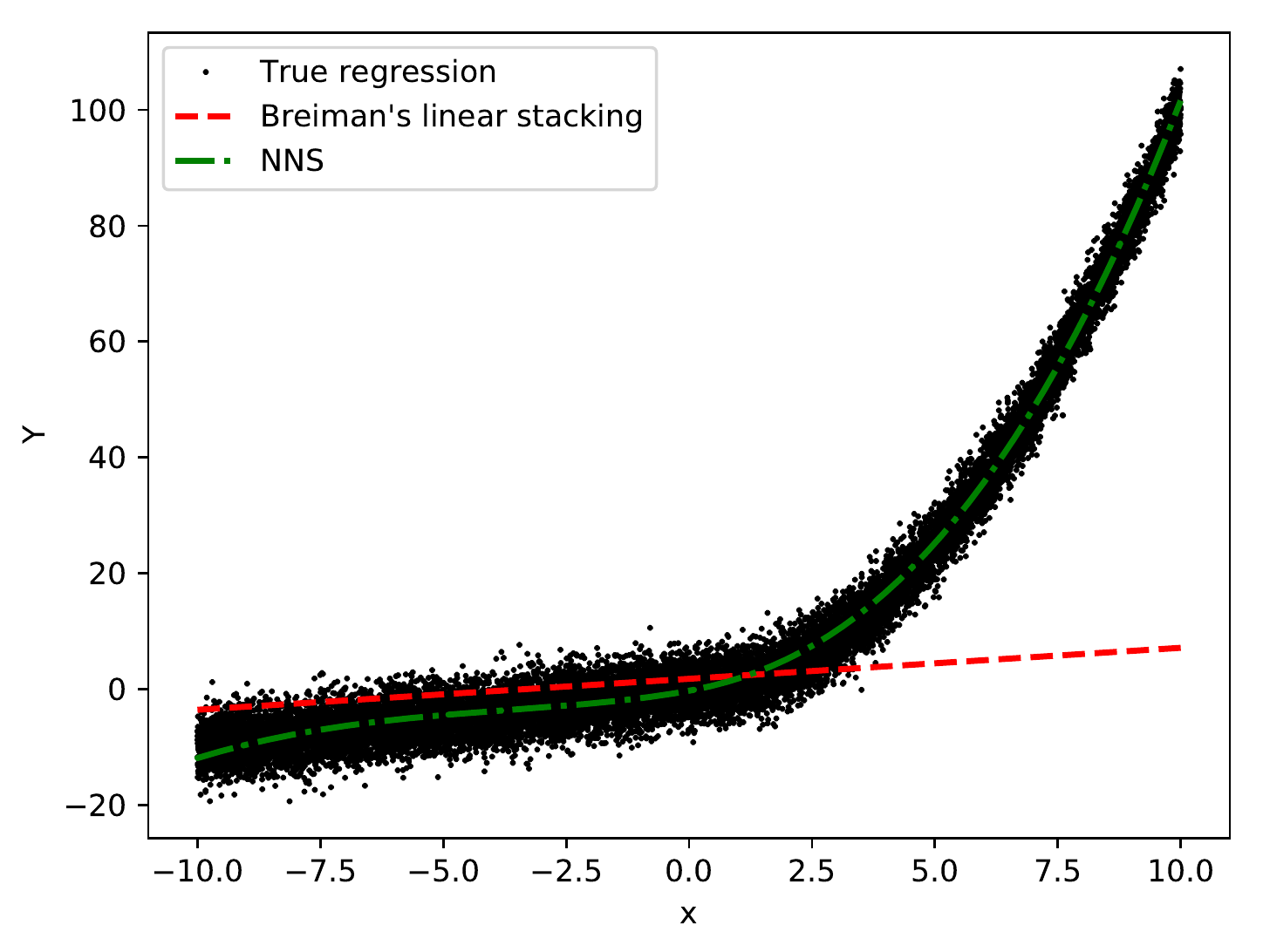}
    \caption{Meta-learners fits for Example \ref{ex:local}.
    While Breiman's meta-learner is not able to fit the true regression satisfactorily, feature varying weights yield better fit.
}
    \label{fig:local2}
\end{figure}
\end{example}

\section{Methodology}
\label{sec:solution}

Our goal is to find
$\theta_\x = (\theta_1(\x),\ldots,\theta_k(\x))'$, 
$\theta_i:\mathcal{X}
\longrightarrow \mathbb{R}$,
 that minimizes the mean squared risk,
\begin{align*}
R(G_\theta)=\E\left[\left(Y-G_\theta(\X)\right)^2\right],
\end{align*}
where
$G_\theta(\x)$ is defined as in Equation in \ref{eq:ours}.

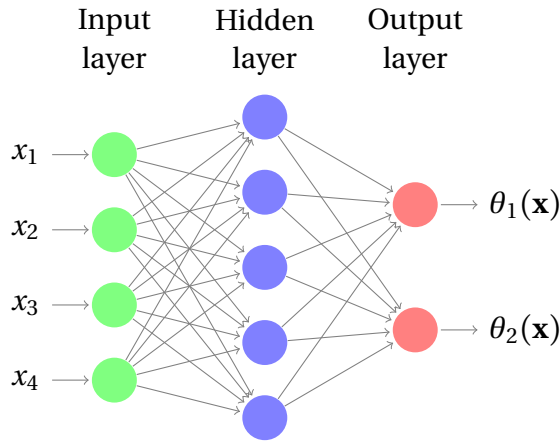
\begin{figure}[!htb]
\def\layersep{2cm}
\centering
\begin{tikzpicture}[shorten >=1pt,->,draw=black!50, node distance=\layersep]
    \tikzstyle{every pin edge}=[<-,shorten <=1pt]
    \tikzstyle{neuron}=[circle,fill=black!25,minimum size=17pt,inner sep=0pt]
    \tikzstyle{input neuron}=[neuron, fill=green!50];
    \tikzstyle{output neuron}=[neuron, fill=red!50];
    \tikzstyle{hidden neuron}=[neuron, fill=blue!50];
    \tikzstyle{annot} = [text width=4em, text centered]

    \foreach \name / \y in {1,...,4}
        \node[input neuron, pin=left:$x_\y$] (I-\name) at (0,-\y) {};

    \foreach \name / \y in {1,...,5}
        \path[yshift=0.5cm]
            node[hidden neuron] (H-\name) at (\layersep,-\y cm) {};

    \foreach \name / \y in {1,...,2}
            \node[output neuron,pin={[pin edge={->}]right:$\theta_{\name}(\x)$}, right of=H-3] (O-\name) at (\layersep,-1.666*\y cm) {};

    \foreach \source in {1,...,4}
        \foreach \dest in {1,...,5}
            \path (I-\source) edge (H-\dest);

    \foreach \source in {1,...,5}
    \foreach \name in {1,...,2}
        \path (H-\source) edge (O-\name);

    \node[annot,above of=H-1, node distance=1cm] (hl) {Hidden layer};
    \node[annot,left of=hl] {Input layer};
    \node[annot,right of=hl] {Output layer};
    \end{tikzpicture}
\caption{Example of a UNNS neural network.}
 \label{fig:unns}
\end{figure}

We  estimate $\theta_\x$ via  an artificial neural network. This network takes $\x$ as input and produces an output $\theta_\x$, which is then used to obtain $G_\theta(\x)$. To estimate the weights of the networks,  we introduce an appropriate loss function that captures the goal of having a  small $R(G_\theta)$. This is done
by using the loss function
\begin{align*}
\frac{1}{n}\sum_{k=1}^n(G_\theta(\x_k)-y_k)^2.    
\end{align*}
Notice that the base regression estimators are  used only when evaluating the loss function; they are not the inputs of the network.
With this approach, we allow each
$\theta_i(\x)$ to be a complex function of the data.
We call this method \emph{Unconstrained Neural Network Stacking} (UNNS). Figure \ref{fig:unns} illustrates  a UNNS that stacks 2 base estimators in a regression problem with four features.

In addition to the linear stacking, this approach allows the user to easily take advantage 
of the neural network architecture by directly adding a network output node, $\phi(\x)$, to the stacking. That is, we also consider
a variation of UNNS which takes the shape
\begin{align*}
    G'_\theta(\x) = \theta'_\x g_\x + \phi(\x).
\end{align*}
This has some similarity to adding a single neural network estimator to the stacking. However, we use the same architecture to create the additional term,  mitigating computation time. 
Algorithm \ref{alg:direct} shows how this method is implemented. In order to avoid over-fitting,
$\theta_i$'s and $g_i$'s are estimated using different folds of the training set.
 
\begin{algorithm}
  \caption{ \small UNNS}
  \label{alg:direct}
  \algorithmicrequire \ {\small  
  Estimation algorithms $g = (g_1, g_2, \ldots, g_k)'$, a dataset $D = (X, Y)$ with $n$ instances (rows), a neural network $N$, features to predict $X^{(p)}$, the amount of folds $F$.
  } \\
  \algorithmicensure \ {\small 
  Predicted values $y^{(p)}$.
  }

  \begin{algorithmic}[1]
    \State Let $I=\{I_o: o \in \{1,2,...,F\}\}$ be a random F-fold partition of the dataset instances, let $I^{(X)}$ refer to a partition of features and $I^{(Y)}$ refer to a partition of the response variable, both being partitioned on the same indices (i.e.: $I_o(i) = (I_o^{(X)}(i), I_o^{(Y)}(i))$ for every $o \in \{1,2,...,F\}\}$ and every $i \in \{1,2,...,n\}\}$), with the partition of indices $\{1, 2, ..., n\}$ represented by $I^{(l)}$.
    \State Let $P$ be a $(n, k)$ matrix.
    \State For $o \in \{1, 2, ..., F\}$ and $j \in \{1, 2, ..., k\}$, fit $g_j$ to $D \backslash I_o$, then use the fitted model to predict $I^{(X)}_o$ and store these predicted values on $P$ (in column $j$ and lines corresponding to $I_o^{(l)})$).
    \State Let $\{g_1^{(f)}, g_2^{(f)}, ..., g_k^{(f)}\}$ be the models $g$ fitted using the whole dataset $D$.
    \State Train the neural network $N$ with each input instance $i$ given by a row of $X$; with $\theta(X_i) = (\theta_1(X_i), \theta_2(X_i), ..., \theta_k(X_i))$ and a scalar $\phi(X_i)$ as outputs; and with loss function given by $(\sum_{j=1}^k \theta_j(X_i) P_{ij} + \phi(X_i) - y_i)^2$ (note: the additional scalar $\phi$ is optional, i.e.: it can be set to zero).
    \State For each instance $i$ of $X^{(p)}$, the corresponding predicted value $Y^{(p)}_i$ is then given by $\sum_{j=1}^k \theta_j(X^{(p)}_i) g_j^{(f)}(X^{(p)}_i) + \phi(X^{(p)}_i)$ where $\theta(X^{(p)}_i)$ and $\phi(X^{(p)}_i)$ are outputs of the neural network (i.e.: $N(X^{(p)}_i)$).
  \end{algorithmic}
\end{algorithm}

In order to achieve
an interpretable
stacked solution, we follow Breiman's suggestion and  consider a second approach to estimate $\theta_i$'s which consists in minimizing $R(G_\theta)$ under the constrain that $\theta_i$'s are weights, that is,  $\theta_i(\x)\geq 0$
and $\sum_{i=1}^k \theta_i(\x)=1$. 
Unfortunately, it is challenging to directly impose this restriction to the solution of the neural network. Instead, we use a different parametrization of the problem, which is motivated by
Theorem \ref{theorem:bestConstraint}.
\begin{theorem}
\label{theorem:bestConstraint}
The solution of 
$$\arg\min_{\theta_x} R(G_\theta)$$
under the constrain that
$\theta_i(\x) \geq 0$
and $\sum_{i=1}^k \theta_i(\x)=1$
is given by
\begin{align}
\label{eq:solutionconstrained}
\theta_\x=\frac{\mathbb{M}^{-1}_\x \vec{e}}{\vec{e}' \mathbb{M}^{-1}_\x \vec{e}},
\end{align}
where $\vec{e}$ is a $k$-dimensional vector of ones and
\begin{align*}
\mathbb{M}_\x &= [\E\left[(Y-g_i(\x)) (Y-g_j(\x)) \right|\X=\x]_{ij}] \\&
= \E[Y^2|\x]-\E[Y|\x](g_i(\x)-g_j(\x)))+g_i(\x)g_j(\x).
\end{align*}
with $(i,j) \in \{1,...,k\}^2$.
\end{theorem}
 
\begin{figure}[!htb]
\def\layersep{2cm}
\centering
\begin{tikzpicture}[shorten >=1pt,->,draw=black!50, node distance=\layersep]
    \tikzstyle{every pin edge}=[<-,shorten <=1pt]
    \tikzstyle{neuron}=[circle,fill=black!25,minimum size=17pt,inner sep=0pt]
    \tikzstyle{input neuron}=[neuron, fill=green!50];
    \tikzstyle{output neuron}=[neuron, fill=red!50];
    \tikzstyle{hidden neuron}=[neuron, fill=blue!50];
    \tikzstyle{annot} = [text width=4em, text centered]

    \foreach \name / \y in {1,...,4}
        \node[input neuron, pin=left:$x_\y$] (I-\name) at (0,-\y) {};

    \foreach \name / \y in {1,...,5}
        \path[yshift=0.5cm]
            node[hidden neuron] (H-\name) at (\layersep,-\y cm) {};

    \node[output neuron,pin={[pin edge={->}]right:$l_{1,1}$}, right of=H-3] (O-1) at (\layersep,-1.25 cm) {};
    \node[output neuron,pin={[pin edge={->}]right:$l_{1,2}$}, right of=H-3] (O-2) at (\layersep,-2.5 cm) {};
    \node[output neuron,pin={[pin edge={->}]right:$l_{2,2}$}, right of=H-3] (O-3) at (\layersep,-3.75 cm) {};

    \foreach \source in {1,...,4}
        \foreach \dest in {1,...,5}
            \path (I-\source) edge (H-\dest);

    \foreach \source in {1,...,5}
    \foreach \name in {1,...,3}
        \path (H-\source) edge (O-\name);

    \node[annot,above of=H-1, node distance=1cm] (hl) {Hidden layer};
    \node[annot,left of=hl] {Input layer};
    \node[annot,right of=hl] {Output layer};
    \end{tikzpicture}
\caption{Example of the CNNS neural network.}
 \label{fig:cnns}
\end{figure}
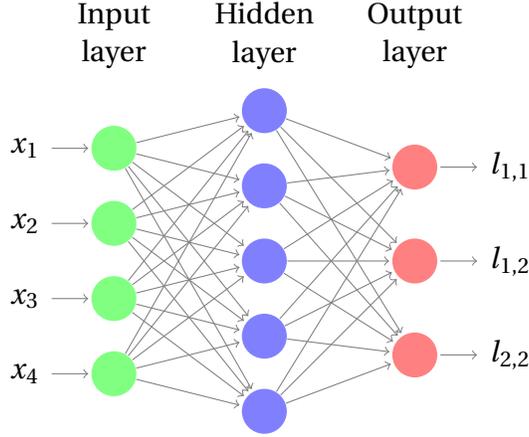

Theorem \ref{theorem:bestConstraint} shows that, under the given constrains, $\theta(\x)$ is uniquely defined by $\mathbb{M}^{-1}_\x$. 
Now, because $\mathbb{M}_\x$ is a covariance matrix,
then  $\mathbb{M}^{-1}_\x$ is positive definite, and thus
Cholesky decomposition
can be applied to it. It follows that $\mathbb{M}^{-1}_\x=L_\x L_\x'$, where $L_\x$ is a lower triangular matrix.
This suggests that we  estimate $\theta_\x$ by first estimating $L_\x$ and then plugging the estimate back into Equation \ref{eq:solutionconstrained}. That is, in order to obtain a good estimator under the above mentioned restrictions, the output of the network is set to be $L_\x$ rather than the weights themselves\footnote{Since the gradients for all matrix operations are implemented for Pytorch tensor classes, the additional operations of the CNNS method will be automatically backpropagated once Pytorch's backward method is called on the loss evaluation.}.  We name this method \emph{Constrained Neural Network Stacking} (CNNS). Figure \ref{fig:cnns} illustrates  a CNNS that stacks 2 base regressors (that is, $L_\x = [l_{ij}]$ is a $2x2$ triangular matrix) in a 4 feature regression problem.

Algorithm \ref{alg:indirect} shows the implementation of this method. As with UNNS, we  also explore a variation which adds an extra network output $\phi(\x)$ to $G_\theta$.

\begin{algorithm}
  \caption{ \small CNNS}
  \label{alg:indirect}
  \algorithmicrequire \ {\small  
  Estimation algorithms $g = (g_1, g_2, \ldots, g_k)'$, a dataset $D = (X, Y)$ with $n$ instances (rows), a neural network $N$, features to predict $X^{(p)}$, the amount of folds $F$.
  } \\
  \algorithmicensure \ {\small 
  Predicted values $y^{(p)}$.
  }

  \begin{algorithmic}[1]
    \State Follow steps 1 to 4 from algorithm \ref{alg:direct}.
    \State Train the neural network $N$ with each input instance $i$ given by a row of $X$;
    with a lower triangular matrix $L_{X_i}$ and a scalar $\phi(X_i)$ as outputs;
    and with loss function given by $(\sum_{j=1}^k \theta_j(X_i) P_{ij} + \phi(X_i) - y_i)^2$,
    where
    $\mathbb{M}^{-1} = L_{X_i} L_{X_i}'$
    ;
    $\theta(X_i)=\frac{\mathbb{M}^{-1}_\x \vec{e}}{\vec{e}' \mathbb{M}^{-1}_\x \vec{e}}$
    and
    $\vec{e}$ is a $k$-dimensional vector of ones
    (note: the additional scalar $\phi$ is optional, i.e.: it can be set to zero).
    \State The predicted value $Y^{(p)}$ is calculated analogously to Algorithm \ref{alg:direct}.
  \end{algorithmic}
\end{algorithm}

\begin{figure}[!htb]
    \centering
    \includegraphics[width=0.9\linewidth]{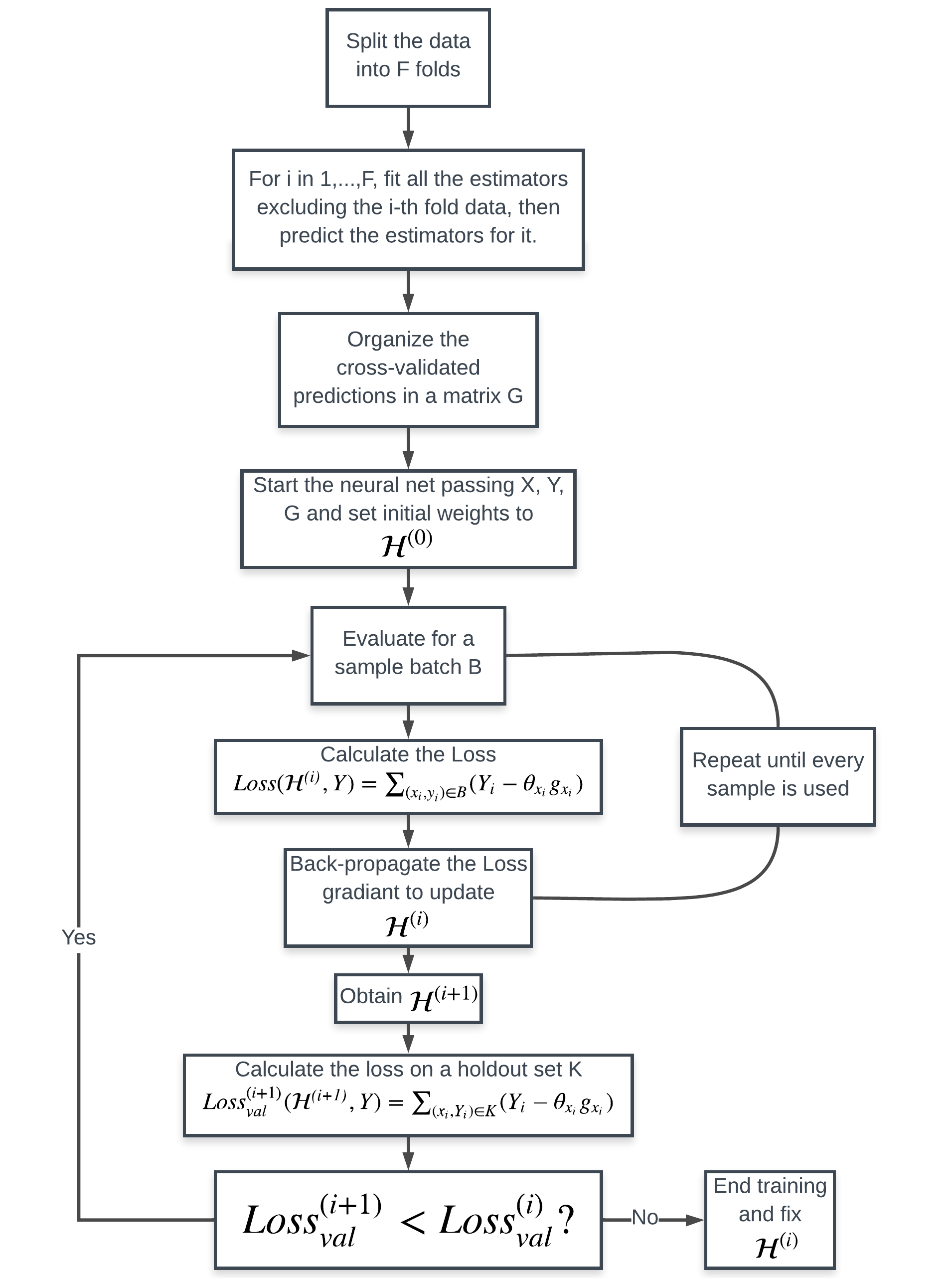}
    \caption{Full NN-Stacking training process.}
    \label{fig:diag}
\end{figure}

Figure \ref{fig:diag} illustrates the full training process. For simplicity, the neural network early stopping patience criterion is set to a single epoch and the additional parameter $\phi_\x$ is not used.

\subsection{Comparison with standard stacking methods}
\label{sec:meta}
Most stacking methods  create a meta-regression model
by applying a regression method directly on the outputs of individual predictions. In particular, a meta-regression method can be a neural network. Such procedure differs from NN-Stacking by the shape of both the input and of the output of the network. While standard stacking uses base regression estimates ($g_x$) as input and $Y$ as output, NN-Stacking uses the features as input and either the weights $\theta_x$ (for UNSS) or $L_\x$ (for CNSS) as outputs. The base regression estimates are  used only on the loss function.
Thus, the NN-Stacking method leads to more interpretable models. Section \ref{sec:app} compares these methods  in terms of their predictive power. We also point out that our approach has some similarity to
\citet{sill2009feature}, which allows each $\theta_i$ to depend on  meta-features computed from $\x$
using a specific parametric form. Neural networks, on the other hand, provide a richer family of functions to model such dependencies (in fact, they are universal approximators; \citet{csaji2001approximation}).

\subsection{Selecting base regressors}

Consider the extreme case where $g_i(\x) = g_j(\x) \, \forall \, \x \in \mathcal{X}$ for some $i \neq j$, that is, the case in which two base regressors generate the same prediction over all  feature space. Now, suppose that one fits a NNS (either CNNS or UNNS) for this case. Then
$    \theta_i g_i(\x) + \theta_j g_j(\x)  =  (\theta_i + \theta_j) g_i(\x).$
Thus, one of the regressions  can be dropped from the stacking with no loss in predictive power.

In practice, our experiments (Section \ref{sec:app})  show that regression estimators that have  strongly correlated results do not contribute to the meta-learner. This suggests that one should choose base regressors with considerably distinct nature.

\subsection{Implementation details}
\label{sec:implementation}
A Python package that implements the methods proposed in this paper is available at \url{github.com/randommm/nnstacking}. The  scripts for the experiments in Section \ref{sec:app} are availiable at \url{github.com/vcoscrato/NNStacking}. 
We work with the following specifications for the artificial neural networks:

\begin{itemize}
\item \textbf{Optimizer}: we use the Adam algorithm \citep{adam-optim} and decrease its learning rate after the validation loss stops improving for a user-defined number of epochs.

\item \textbf{Initialization}:  we use the Xavier Gaussian method proposed by \citet{nn-initialization} to sample the initial parameters of the neural network.

\item \textbf{Layer activation and regularization}: we use ELU \citep{elu} as the activation function, and do not use  regularization.

\item \textbf{Normalization}: we use batch normalization \citep{batch-normalization}   to speed-up the training process.

\item \textbf{Stopping criterion}: in order to address the risk of having strong over-fit on the neural networks, we worked with a 90\%/10\% split early stopping for small datasets and a higher split factor for larger datasets (increasing the proportion of training instances) and a patience of 10 epochs without improvement on the validation set.

\item \textbf{Dropout}:  
We use dropout (with a rate of 50\%)
to address the problem of over-fitting \citep{dropout}.

\item \textbf{Software}: we use PyTorch \citep{pytorch}. 

\item \textbf{Architecture}: as default values we use a 3 layer depth network with hidden layer size set to 100; these values have been experimentally found to be suitable in our experiments (Section \ref{sec:app}).

\end{itemize}

\section{Experiments}
\label{sec:app}

We compare stacking methods for the following  UCI datasets:
\begin{itemize}
    \item The GPU kernel performance dataset (241600 instances, 13 features) \citep{nugteren2015cltune},
    \item The music year prediction dataset \citep{Dua:2017} (515345 instances and 90 features),
    \item The blog feedback dataset \citep{buza2014feedback} (60021 instances, 280 features),
    \item The superconductivity dataset \citep{hamidieh2018data} (21263 instances, 80 features).
\end{itemize}

First, 
we fit the following regression estimators (that will be stacked):
\begin{itemize}
    \item Three linear models: with L1, L2, and no penalization \citep{friedman2001elements},
    \item Two tree based models: bagging and random forests \citep{friedman2001elements},
    \item A gradient boosting method (GBR) \citep{meir2003introduction}.
\end{itemize}
The tuning
parameters of these estimators are chosen by
cross-validation using scikit-learn \citep{scikit-learn}.

Using these base estimators, we then fit four variations of NNS (both CNNS and UNNS with and without the additional $\phi_x$) using the following specifications:
\begin{itemize}
\item \textbf{Tuning}: three different architectures were tested for each neural network approach. The layer size was fixed at 100 and the number of hidden layers were set to 1, 3, and 10. We choose the architecture with the lowest validation mean-squared error.

\item \textbf{Train/validation/test split}: for all  datasets, we use 75\% of the instances to fit the models, among which
 10\%
are used for performing early stop.
The remaining 25\% of the instances are used as a test set to compare the performance of the various models.
The train/test split is performed at random. The cross-validated predictions (the matrix $P$ denoted on Algorithm \ref{alg:direct}) are obtained using a 10-fold cross-validation on the training data (i.e., $F = 10$).

\item \textbf{Total fitting time}: we compute the total fitting time
(in seconds; including the time for cross-validating the  network architecture) of each  method on two cores of an AMD Ryzen 7 1800X processor running at 3.6Gz. 

\end{itemize}

We compare our methods with  Breiman's linear stacking and the usual neural net stacking model described in Section \ref{sec:meta}. In addition to these, we also include a comparison with a direct neural network that has $\x$ as its input and $Y$ as its output.
 
The comparisons are made by evaluating the mean squared error (MSE, $n^{-1}\sum_{i=1}^n (y_i-g(\x_i))^2$) and the mean absolute error (MAE, $n^{-1}\sum_{i=1}^n |y_i-g(\x_i)|$) of each model $g$ on a test set. We also compute the standard error for each of these metrics, which enables one to compute  confidence intervals for the errors of each method.


\subsection{GPU kernel performance dataset}
\label{sec:GPU}

Table \ref{tab:GPU_err} shows the results that were obtained  for the GPU kernel performance dataset. Our UNNS methods  outperforms both Breiman's stacking and the usual meta-regression stacking approaches in terms of MSE. Moreover, the UNNS model is also the best one in terms of MAE, even though the gap between the models is lower in this case. Our stacking methods also perform better than all base estimators. This suggests that each base model performs better on a distinct region of the feature space.

\begin{figure}[!htb]
    \centering
    \includegraphics[scale=0.8]{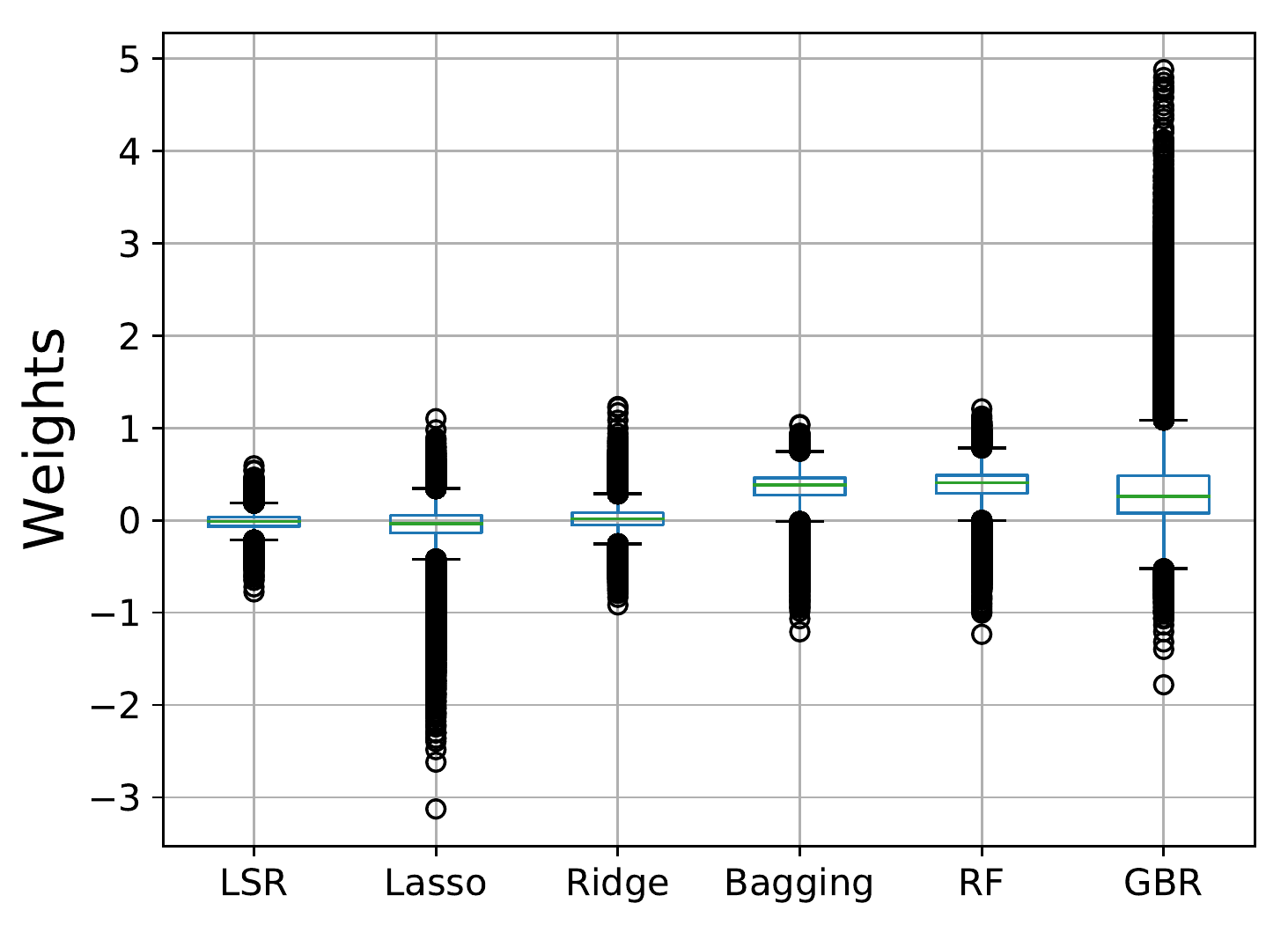}
    \caption{Weight distribution for the GPU kernel performance dataset.}
    \label{fig:GPU_weights}
\end{figure}

Figure \ref{fig:GPU_weights} shows a boxplot with the distribution of the  fitted $\theta_i$'s for UNNS.
Many fitted values fall out of the range $[0, 1]$, which explains why UNNS gives better results than Breiman's and CNNS (which have the restriction that $\theta_i$'s must be proper weights).

Table \ref{tab:GPU_cor} shows the correlation between the prediction errors for base estimators. The linear estimators had an almost perfect  pairwise correlation, which indicates that removing up to 2 of them from the stacking would not affect predictions. Indeed, after refitting UNNS without using ridge regression and lasso, we obtain exactly the same results. We also refit the best UNNS removing all of the linear estimators to check if poor performing estimators are making stacking results worse. In this setting, we obtain an  MSE of $11074.13 (\pm 227.57)$, and a MAE of $45.76 (\pm 0.39)$. Note that although the point estimates of the errors are lower than those obtained in Table \ref{tab:GPU_err}, the confidence intervals have an intersection, which leads to the conclusion that the poor performance of linear estimators is  not damaging  the stacked estimator. 

\begin{table}[!htb]
\centering
\begin{adjustbox}{max width=\textwidth}
\begin{tabular}{|c|l|l|l|l|}
\hline
\multicolumn{1}{|l|}{Type}                                                    & Model                                  & MSE                      & MAE                & Total fit time \\ \hline
\multirow{6}{*}{\begin{tabular}[c]{@{}c@{}}Stacked\\ estimators\end{tabular}} & UNNS + $\phi_x$ (3 layers)             & 11400.43 ($\pm$ 250.03)  & 45.91 ($\pm$ 0.39) & 3604           \\ \cline{2-5} 
                                                                              & CNNS + $\phi_x$ (3 layers)             & 19371.98 ($\pm$ 429.96)  & 53.09 ($\pm$ 0.52) & 3531           \\ \cline{2-5} 
                                                                              & UNNS (3 layers)                        & 11335.85 ($\pm$ 241.94)  & 45.85 ($\pm$ 0.39) & 3540           \\ \cline{2-5} 
                                                                              & CNNS (3 layers)                        & 18748.66 ($\pm$ 424.5)   & 51.65 ($\pm$ 0.52) & 3387           \\ \cline{2-5} 
                                                                              & Breiman's stacking                     & 30829.11 ($\pm$ 717.13)  & 62.41 ($\pm$ 0.67) & 63             \\ \cline{2-5} 
                                                                              & Meta-regression neural net (10 layers) & 24186.4 ($\pm$ 545.52)   & 58.79 ($\pm$ 0.59) & 85             \\ \hline
\begin{tabular}[c]{@{}c@{}}Direct\\ estimator\end{tabular}                    & Direct neural net (10 layers)          & 14595.98 ($\pm$ 307.11)  & 52.3 ($\pm$ 0.44)  & 380            \\ \hline
\multirow{6}{*}{\begin{tabular}[c]{@{}c@{}}Base\\ estimators\end{tabular}}    & Least squares                          & 79999.09 ($\pm$ 1504.75) & 176.41 ($\pm$ 0.9) & -              \\ \cline{2-5} 
                                                                              & Lasso                                  & 80091.85 ($\pm$ 1526.05) & 175.5 ($\pm$ 0.9)  & -              \\ \cline{2-5} 
                                                                              & Ridge                                  & 79999.05 ($\pm$ 1504.76) & 176.41 ($\pm$ 0.9) & -              \\ \cline{2-5} 
                                                                              & Bagging                                & 31136.93 ($\pm$ 737.47)  & 62.35 ($\pm$ 0.67) & -              \\ \cline{2-5} 
                                                                              & Random forest                          & 30923.64 ($\pm$ 727.99)  & 62.2 ($\pm$ 0.67)  & -              \\ \cline{2-5} 
                                                                              & Gradient boosting                      & 32043.23 ($\pm$ 676.1)   & 90.51 ($\pm$ 0.63) & -              \\ \hline
\end{tabular}
\end{adjustbox}
\caption{Evaluation of model accuracy metrics for the GPU kernel performance dataset.}
\label{tab:GPU_err}
\end{table}

\begin{table}[!htb]
\centering
\begin{adjustbox}{max width=\textwidth}
\begin{tabular}{|c|c|c|c|c|c|c|} \hline
Models &  Least squares &  Lasso &  Ridge &  Bagging &  Random forest &  Gradient boosting \\ \hline
Least squares     &           1.00 &   1.00 &   1.00 &     0.39 &           0.39 &               0.80 \\ \hline
Lasso             &           1.00 &   1.00 &   1.00 &     0.39 &           0.39 &               0.80 \\ \hline
Ridge             &           1.00 &   1.00 &   1.00 &     0.39 &           0.39 &               0.80 \\ \hline
Bagging           &           0.39 &   0.39 &   0.39 &     1.00 &           0.98 &               0.62 \\ \hline
Random forest     &           0.39 &   0.39 &   0.39 &     0.98 &           1.00 &               0.62 \\ \hline
Gradient boosting &           0.80 &   0.80 &   0.80 &     0.62 &           0.62 &               1.00 \\ \hline
\end{tabular}
\end{adjustbox}
\caption{Pearson correlation between base estimators prediction errors for the GPU kernel performance dataset.}
\label{tab:GPU_cor}
\end{table}

\subsection{Music year dataset}
\label{sec:music}

Table \ref{tab:music_err} shows the accuracy metrics results for the music year dataset. In this case, the CNNS gave the best results, both in terms of MSE and MAE.  For this dataset,  Breiman's stacking was  worse than using gradient boosting, one of the base regressors. The same happens with the usual meta-regression neural network approach. On the other hand, NNS could find a representation
that combines the already powerful GBR estimator with less powerful ones in a way that leverages their individual performance.

All base estimators had high prediction error correlations (Table \ref{tab:music_cor}). In particular, two of the linear estimators could be removed from the stacking without affecting its performance. However, when removing all three linear estimators
the MSE for the best NNS  increased to $83.92 (\pm 0.57)$ and its MAE increased to $6.44 (\pm 0.02)$. 

\begin{figure}[!htb]
    \centering
    \includegraphics[scale=0.8]{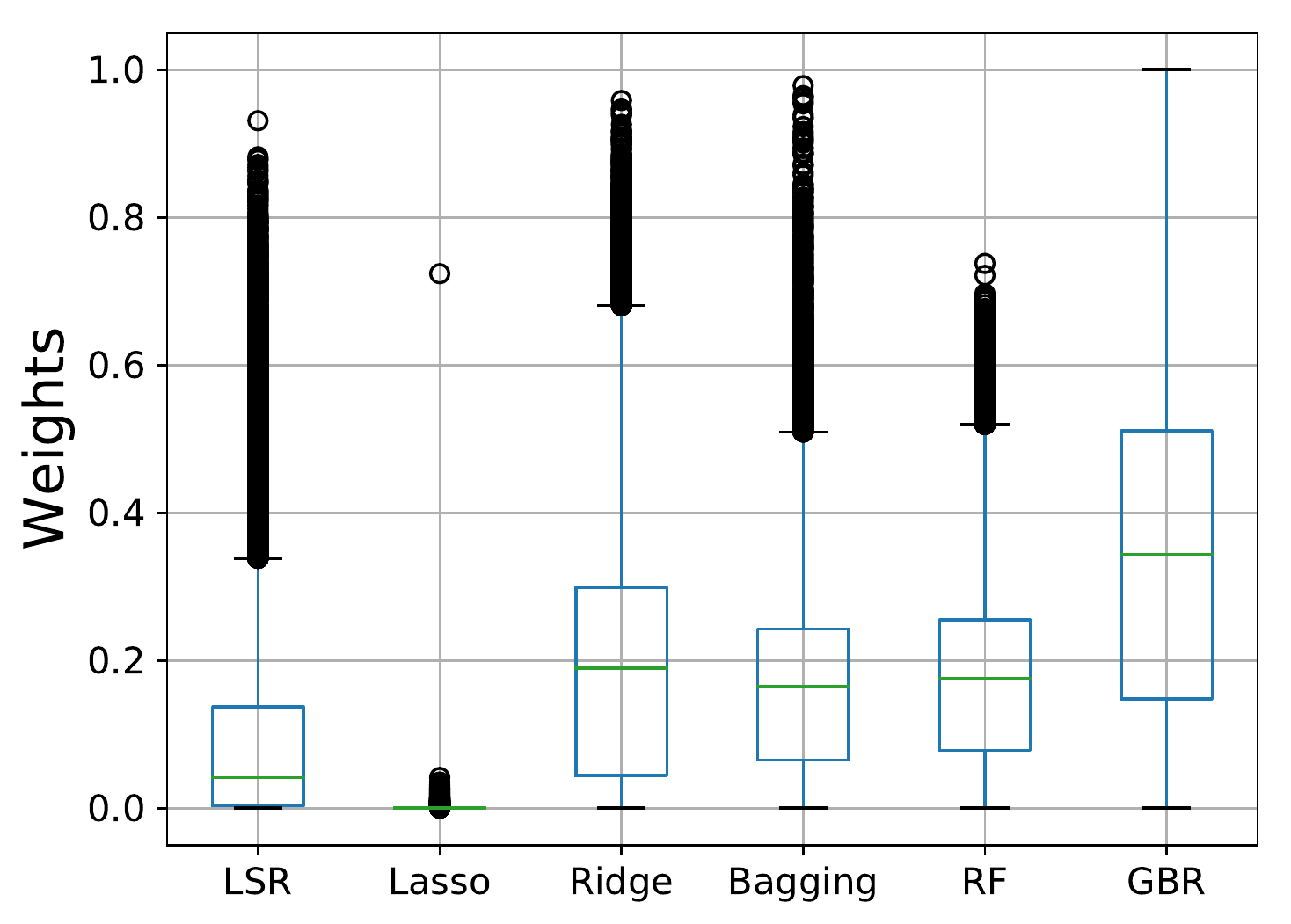}
    \caption{Weight distribution for the music year dataset.}
    \label{fig:music_weights}
\end{figure}

Figure \ref{fig:music_weights} shows that the fitted NNS weights have a large dispersion. This illustrates the flexibility added by our method. Models with very distinctive nature (e.g., ridge regression - which imposes a linear shape on the regression function, and random forests - which is fully non-parametric) can add to each other, getting weights of different magnitudes depending on the region of the feature space that the new instance lies on.

\begin{table}[!htb]
\centering
\begin{adjustbox}{max width=\textwidth}
\begin{tabular}{|c|l|l|l|l|}
\hline
\multicolumn{1}{|l|}{Type}                                                    & Model (Best architecture)                        & MSE                  & MAE                & Total fit time \\ \hline
\multirow{7}{*}{\begin{tabular}[c]{@{}c@{}}Stacked\\ estimators\end{tabular}} & UNNS + $\phi_\x$ (10 layers)          & 92.37 ($\pm$ 7.18)   & 6.53 ($\pm$ 0.02)  & 9432           \\ \cline{2-5} 
                                                                              & CNNS + $\phi_\x$ (3 layers)           & 83.05 ($\pm$ 0.57)   & 6.38 ($\pm$ 0.02)  & 8851           \\ \cline{2-5} 
                                                                              & UNNS (10 layers)                     & 95.35 ($\pm$ 1.81)   & 7.45 ($\pm$ 0.02)  & 12087          \\ \cline{2-5} 
                                                                              & CNNS (3 layers)                      & 82.99 ($\pm$ 0.57)   & 6.38 ($\pm$ 0.02)  & 11466          \\ \cline{2-5} 
                                                                              & Breiman's stacking                   & 87.66 ($\pm$ 0.57)   & 6.61 ($\pm$ 0.02)  & 3090           \\ \cline{2-5} 
                                                                              & Meta-regression neural net (1 layer) & 87.64 ($\pm$ 0.59) & 6.61 ($\pm$ 0.02)  & 571              \\ \hline
\multicolumn{1}{|c|}{\begin{tabular}[c]{@{}c@{}}Direct\\ estimator\end{tabular}} & Direct neural net (1 layer)          & 1596.2 ($\pm$ 10.88) & 29.83 ($\pm$ 0.07) & 2341           \\ \hline
\multirow{6}{*}{\begin{tabular}[c]{@{}c@{}}Base\\ estimators\end{tabular}}    & Least squares                        & 92.03 ($\pm$ 0.62)   & 6.82 ($\pm$ 0.02)  & -              \\ \cline{2-5} 
                                                                              & Lasso                                & 92.61 ($\pm$ 0.62)   & 6.87 ($\pm$ 0.02)  & -              \\ \cline{2-5} 
                                                                              & Ridge                                & 92.03 ($\pm$ 0.62)   & 6.82 ($\pm$ 0.02)  & -              \\ \cline{2-5} 
                                                                              & Bagging                              & 92.83 ($\pm$ 0.59)   & 6.84 ($\pm$ 0.02)  & -              \\ \cline{2-5} 
                                                                              & Random forest                        & 92.6 ($\pm$ 0.59)    & 6.83 ($\pm$ 0.02)  & -              \\ \cline{2-5} 
                                                                              & Gradient boosting                    & 87.49 ($\pm$ 0.6)    & 6.58 ($\pm$ 0.02)  & -              \\ \hline
\end{tabular}
\end{adjustbox}
\caption{Evaluation of the model accuracy metrics for the music year dataset.}
\label{tab:music_err}
\end{table}

\begin{table}[!htb]
\centering
\begin{adjustbox}{max width=\textwidth}
\begin{tabular}{|c|c|c|c|c|c|c|} \hline
Models &  Least squares &  Lasso &  Ridge &  Bagging &  Random forest &  Gradient boosting \\ \hline
Least squares     &           1.00 &   1.00 &   1.00 &     0.87 &           0.87 &               0.95 \\ \hline
Lasso             &           1.00 &   1.00 &   1.00 &     0.88 &           0.88 &               0.96 \\ \hline
Ridge             &           1.00 &   1.00 &   1.00 &     0.87 &           0.87 &               0.95 \\ \hline
Bagging           &           0.87 &   0.88 &   0.87 &     1.00 &           0.89 &               0.91 \\ \hline
Random forest     &           0.87 &   0.88 &   0.87 &     0.89 &           1.00 &               0.91 \\ \hline
Gradient boosting &           0.95 &   0.96 &   0.95 &     0.91 &           0.91 &               1.00 \\ \hline
\end{tabular}
\end{adjustbox}
\caption{Pearson correlation between base estimators prediction errors for the music year dataset.}
\label{tab:music_cor}
\end{table}

\subsection{Blog feedback dataset}
\label{sec:blog}

Table \ref{tab:blog_err} shows the  results for the blog feedback dataset. All  stacked estimators had similar performance in terms of MSE. However, UNNS had slightly worse performance with respect to MAE. This may happen because the NNS is designed to minimize the MSE and not the MAE. Overall, for this small dataset, the NNS shows no improvement over Breiman's stacking or the usual meta-regression neural network.

\begin{figure}[!htb]
    \centering
    \includegraphics[scale=0.8]{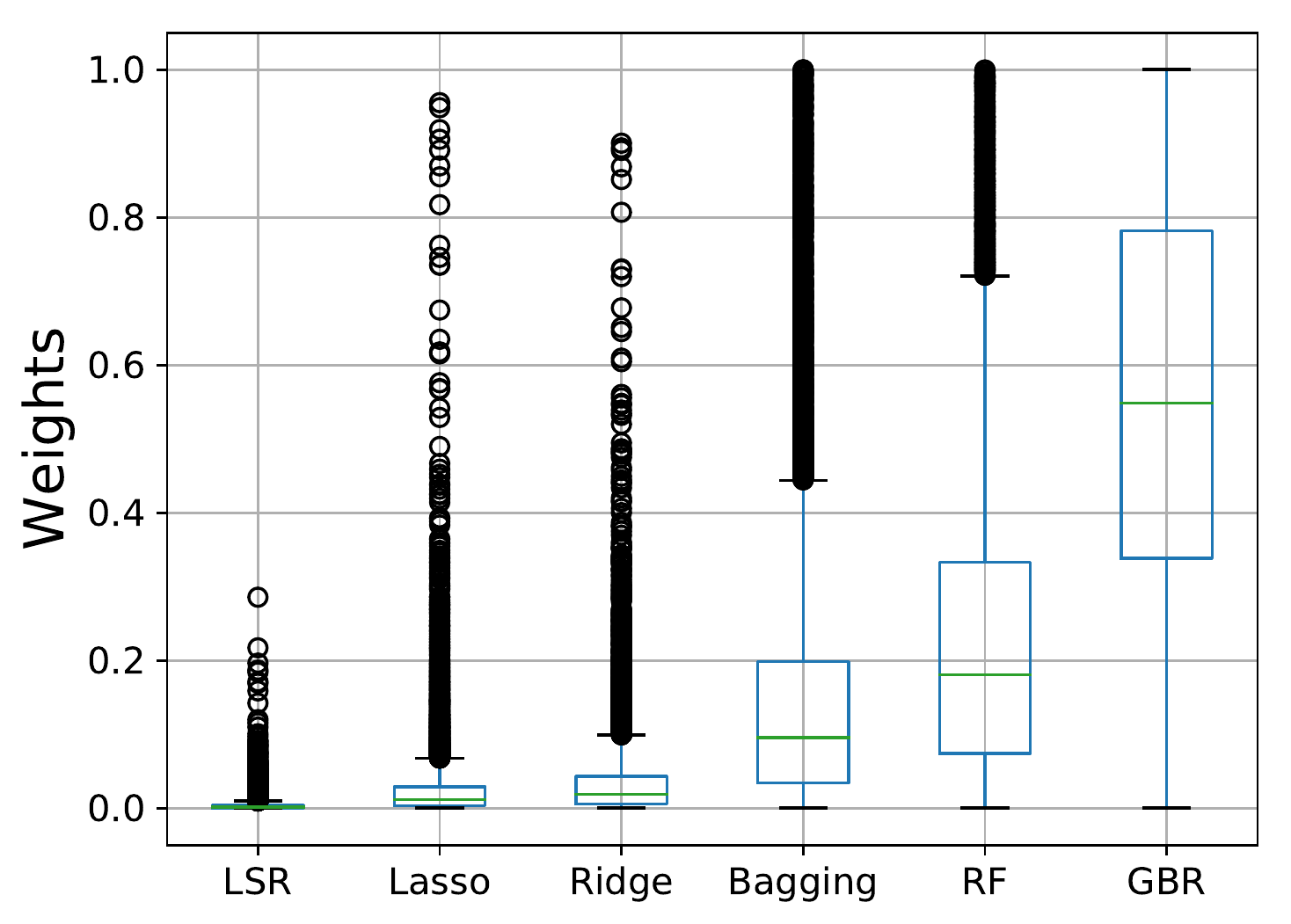}
    \caption{Weight distribution for the blog feedback dataset.}
    \label{fig:blog_weights}
\end{figure}

 GBR had the lowest MSE for the base estimators, while bagging and random forests had the lowest MAE. This explains why these models have larger  fitted weights (Figure \ref{fig:blog_weights}). Moreover, the linear models prediction errors  had an almost perfect error correlation (Table \ref{tab:blog_cor}). This suggests that removing up to 2 of them from the NNS would not impact its performance. Also, the linear estimators 
 has a poor performance when 
 compared to the other base regressors. 
 We thus refit the best NNS for this data after removing these estimators, and achieve an MSE of 531.88 ($\pm 62.67$)  and a MAE of 5.31 ($\pm 0.20$).
 We conclude that the linear estimators
 did not damage nor improved the NNS. 

\begin{table}[!htb]
\centering
\begin{adjustbox}{max width=\textwidth}
\begin{tabular}{|l|l|l|l|l|}
\hline
Type                                                                          & Model (Best architecture)                         & MSE                   & MAE               & Total fit time \\ \hline
\multirow{7}{*}{\begin{tabular}[c]{@{}c@{}}Stacked\\ estimators\end{tabular}} & UNNS + $\phi_\x$ (10 layers)           & 542.02 ($\pm$ 62.65)  & 5.89 ($\pm$ 0.2)  & 420            \\ \cline{2-5} 
                                                                              & CNNS + $\phi_\x$ (1 layer)             & 548.99 ($\pm$ 63.9)   & 5.44 ($\pm$ 0.2)  & 404            \\ \cline{2-5} 
                                                                              & UNNS (10 layers)                      & 557.95 ($\pm$ 61.51)  & 6.38 ($\pm$ 0.2)  & 447            \\ \cline{2-5} 
                                                                              & CNNS (3 layers)                      & 540.68 ($\pm$ 63.87)  & 5.44 ($\pm$ 0.2)  & 433            \\ \cline{2-5} 
                                                                              & Breiman's stacking                    & 593.74 ($\pm$ 73.19)  & 5.41 ($\pm$ 0.21) & 202            \\ \cline{2-5} 
                                                                              & Meta-regression neural net (3 layers) & 537.66 ($\pm$ 63.31)  & 5.53 ($\pm$ 0.2)  & 44             \\ \hline 
\multicolumn{1}{|c|}{\begin{tabular}[c]{@{}c@{}}Direct\\ estimator\end{tabular}} & Direct neural net (3 layers)          & 676.79 ($\pm$ 81.0)   & 7.52 ($\pm$ 0.22) & 63             \\ \hline
\multirow{6}{*}{\begin{tabular}[c]{@{}c@{}}Base\\ estimators\end{tabular}}    & Least squares                         & 878.88 ($\pm$ 109.42) & 9.56 ($\pm$ 0.25) & -              \\ \cline{2-5} 
                                                                              & Lasso                                 & 877.11 ($\pm$ 108.11) & 9.04 ($\pm$ 0.25) & -              \\ \cline{2-5} 
                                                                              & Ridge                                 & 877.92 ($\pm$ 109.47) & 9.53 ($\pm$ 0.25) & -              \\ \cline{2-5} 
                                                                              & Bagging                               & 619.04 ($\pm$ 88.49)  & 5.27 ($\pm$ 0.21) & -              \\ \cline{2-5} 
                                                                              & Random forest                         & 585.22 ($\pm$ 64.88)  & 5.37 ($\pm$ 0.21) & -              \\ \cline{2-5} 
                                                                              & Gradient boosting                     & 557.28 ($\pm$ 63.88)  & 5.75 ($\pm$ 0.2)  & -              \\ \hline
\end{tabular}
\end{adjustbox}
\caption{Evaluation of model accuracy metrics for the blog feedback dataset.}
\label{tab:blog_err}
\end{table}

\begin{table}[!htb]
\centering
\begin{adjustbox}{max width=\textwidth}
\begin{tabular}{|c|c|c|c|c|c|c|} \hline
Models &  Least squares &  Lasso &  Ridge &  Bagging &  Random forest &  Gradient boosting \\ \hline
Least squares     &           1.00 &   0.99 &   1.00 &     0.68 &           0.70 &               0.81 \\ \hline
Lasso             &           0.99 &   1.00 &   0.99 &     0.68 &           0.69 &               0.81 \\ \hline
Ridge             &           1.00 &   0.99 &   1.00 &     0.68 &           0.70 &               0.81 \\ \hline
Bagging           &           0.68 &   0.68 &   0.68 &     1.00 &           0.92 &               0.89 \\ \hline
Random forest     &           0.70 &   0.69 &   0.70 &     0.92 &           1.00 &               0.90 \\ \hline
Gradient boosting &           0.81 &   0.81 &   0.81 &     0.89 &           0.90 &               1.00 \\ \hline
\end{tabular}
\end{adjustbox}
\caption{Pearson correlation between base estimators prediction errors for the blog feedback dataset.}
\label{tab:blog_cor}
\end{table}

\subsection{Superconductivity dataset}
\label{sec:super}

\begin{figure}[!htb]
    \centering
    \includegraphics[scale=0.8]{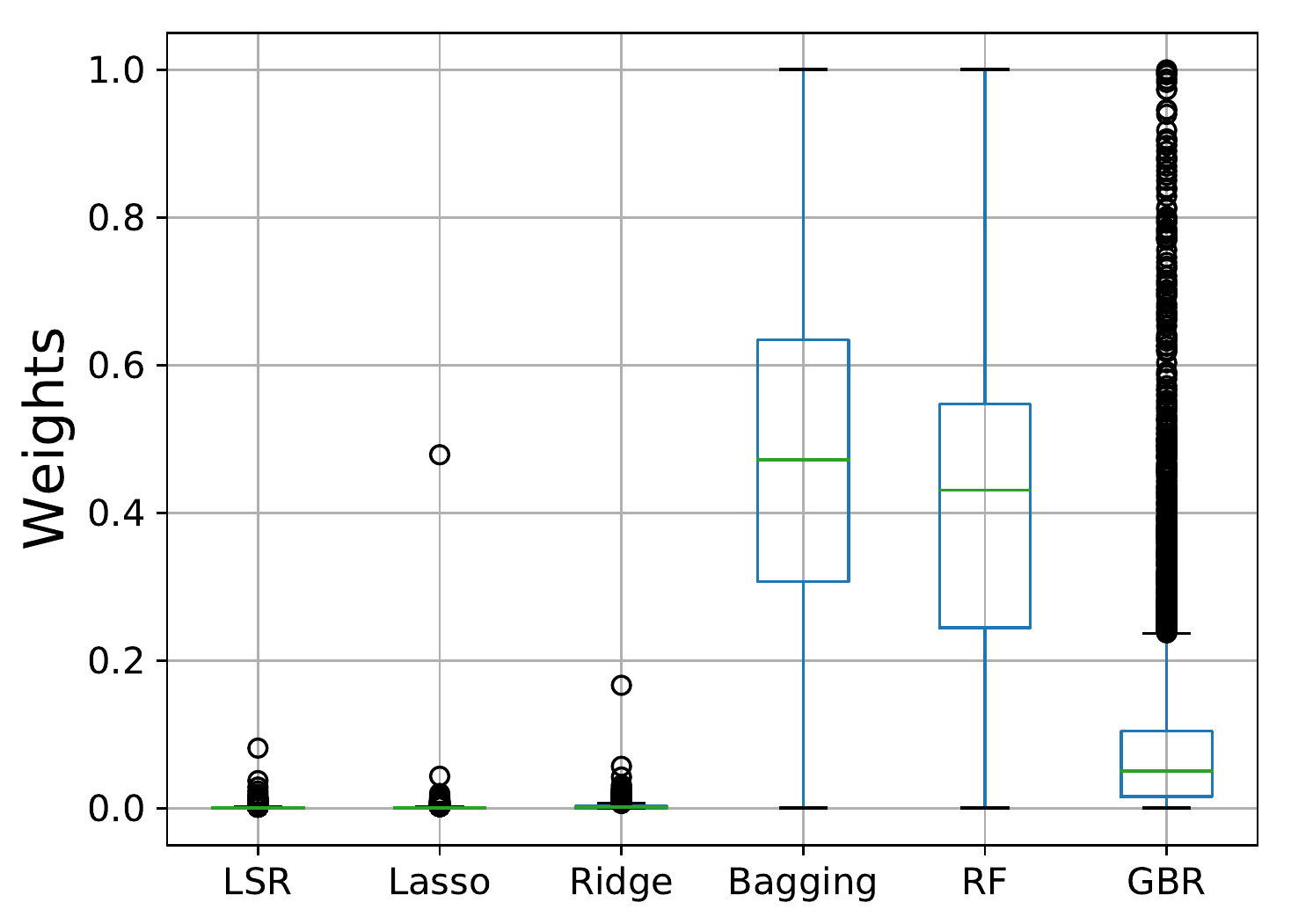}
    \caption{Weight distribution for the superconductivity dataset.}
    \label{fig:super_weights}
\end{figure}

The results for the superconductivity dataset (Table \ref{tab:super_err}) were similar to those obtained for the blog feedback data: the NNS methods perform slightly better than Breiman's in terms of MSE, and worse in terms of MAE. Moreover, both tree-based models had the best MSE among base estimators, competing with the GBR in terms of MAE. Hence, they got larger fitted weights (Figure \ref{fig:super_weights}). 

Table \ref{tab:super_cor} shows that GBR did not have a high correlation error to the tree-based estimators (0.72 in both cases). This is another reason why although having higher MSE, the GBR has high fitted weight for some instances. One can also note that bagging and random forest had an almost perfect error correlation. This implies that 
removing one of them would 
lead to no changes in the NNS. Finally, removing the linear models did not change the MSE and the MAE for the stacking methods.

\begin{table}[!htb]
\centering
\begin{adjustbox}{max width=\textwidth}
\begin{tabular}{|l|l|l|l|l|}
\hline
Type                                                                                                & Model (Best architecture)                        & MSE                  & MAE               & Total fit time \\ \hline
\multicolumn{1}{|c|}{\multirow{7}{*}{\begin{tabular}[c]{@{}c@{}}Stacked\\ estimators\end{tabular}}} & UNNS + $\phi_\x$ (10 layers)         & 98.97 ($\pm$ 4.67)   & 5.71 ($\pm$ 0.11) & 334            \\ \cline{2-5} 
\multicolumn{1}{|c|}{}                                                                              & CNNS + $\phi_\x$ (1 layer)           & 98.79 ($\pm$ 4.67)   & 5.65 ($\pm$ 0.11) & 325            \\ \cline{2-5} 
\multicolumn{1}{|c|}{}                                                                              & UNNS (10 layers)                     & 98.62 ($\pm$ 4.77)   & 5.64 ($\pm$ 0.11) & 344            \\ \cline{2-5} 
\multicolumn{1}{|c|}{}                                                                              & CNNS (3 layers)                      & 98.60 ($\pm$ 4.75)   & 5.60 ($\pm$ 0.11) & 335            \\ \cline{2-5} 
\multicolumn{1}{|c|}{}                                                                              & Breiman's stacking                   & 99.79 ($\pm$ 4.95)   & 5.48 ($\pm$ 0.11) & 48             \\ \cline{2-5} 
\multicolumn{1}{|c|}{}                                                                              & Meta-regression neural net (1 layer) & 99.05 ($\pm$ 4.78)   & 5.60 ($\pm$ 0.11) & 24             \\ \hline
\multicolumn{1}{|c|}{\begin{tabular}[c]{@{}c@{}}Direct\\ estimator\end{tabular}}                                                                              & Direct neural net (3 layers)         & 274.93 ($\pm$ 7.20)  & 7.20 ($\pm$ 0.16) & 62             \\ \hline
\multirow{6}{*}{\begin{tabular}[c]{@{}c@{}}Base\\ estimators\end{tabular}}                          & Least squares                        & 308.65 ($\pm$ 13.41) & 7.12 ($\pm$ 0.16) & -              \\ \cline{2-5} 
                                                                                                    & Lasso                                & 475.6 ($\pm$ 17.08)  & 9.41 ($\pm$ 0.19) & -              \\ \cline{2-5} 
                                                                                                    & Ridge                                & 309.17 ($\pm$ 13.42) & 7.17 ($\pm$ 0.16) & -              \\ \cline{2-5} 
                                                                                                    & Bagging                              & 105.14 ($\pm$ 5.68)  & 5.02 ($\pm$ 0.12) & -              \\ \cline{2-5} 
                                                                                                    & Random forest                        & 103.02 ($\pm$ 5.59)  & 5.08 ($\pm$ 0.12) & -              \\ \cline{2-5} 
                                                                                                    & Gradient boosting                    & 161.48 ($\pm$ 8.74)  & 5.05 ($\pm$ 0.13) & -              \\ \hline
\end{tabular}
\end{adjustbox}
\caption{Evaluation of model accuracy metrics for the superconductivity dataset.}
\label{tab:super_err}
\end{table}

\begin{table}[!htb]
\centering
\begin{adjustbox}{max width=\textwidth}
\begin{tabular}{|c|c|c|c|c|c|c|} \hline
Models &  Least squares &  Lasso &  Ridge &  Bagging &  Random forest &  Gradient boosting \\ \hline
Least squares     &           1.00 &   0.80 &   1.00 &     0.52 &           0.51 &               0.78 \\ \hline
Lasso             &           0.80 &   1.00 &   0.80 &     0.45 &           0.44 &               0.67 \\ \hline
Ridge             &           1.00 &   0.80 &   1.00 &     0.52 &           0.51 &               0.78 \\ \hline
Bagging           &           0.52 &   0.45 &   0.52 &     1.00 &           0.91 &               0.72 \\ \hline
Random forest     &           0.51 &   0.44 &   0.51 &     0.91 &           1.00 &               0.72 \\ \hline
Gradient boosting &           0.78 &   0.67 &   0.78 &     0.72 &           0.72 &               1.00 \\ \hline
\end{tabular}
\end{adjustbox}
\caption{Pearson correlation between base estimators prediction errors for the superconductivity dataset.}
\label{tab:super_cor}
\end{table}

\section{Conclusion and future extensions}
\label{sec:conclusion}

NN-Stacking is a stacking tool with good predictive power that keeps the simplicity in interpretation of Breiman's method. The key idea of the method is to take advantage of the fact that distinct base models often perform better at different regions of the feature space, and thus it allows the weight associated to each model  to vary with $\x$.

Our experiments show  that both CNNS and UNNS can be suitable in different settings: in cases where the base estimators do not capture the complexity from the whole data, the freedom adopted by  UNNS can lead to a larger improvement in performance. On the other hand, when base estimators already have high performance, UNNS the CNNS have similar predictive power, but  the restrictions imposed by  CNNS  guarantee a more 
interpretable solution. 
Both CNNS and UNNS have comparable computational cost.

In our experiments, we observe
that NNS improves over standard stacking approaches especially on large datasets.
This can be explained by the fact that NNS methods have a higher complexity (i.e., larger number of parameters) than the other approaches. Thus, a larger sample size is needed to satisfactorily estimate them. 
The experiments also show that including weak regression methods (such as linear methods)
might decrease the errors of NNS.
In a few cases, however, adding such weak regressors
slightly increases the prediction errors of the stacked estimators This suggests that  adding a penalization to the loss function that  encourages $\theta_i$'s to be zero  may lead to improved results. 



Future work includes
extending these ideas to classification problems, as well as developing a leave-one-out version based on super learners \citep{van2007super}. Also, we desire to develop a method of regularization on population moments estimation to avoid over-fitting, as well as to study asymptotic properties for the estimator of $L_\x$.

\section*{Acknowledgments}
Victor Coscrato and Marco Inácio are grateful for the financial support of CAPES: this study was financed in part by the Coordenação de
Aperfeiçoamento de Pessoal de Nível Superior - Brasil (CAPES) -
Finance Code 001. Rafael Izbicki is grateful for the financial support of FAPESP (grant 2017/03363-8) and CNPq (grant 306943/2017-4).
The authors are also grateful for the suggestions
given by Rafael Bassi Stern.

\section*{References}
\bibliography{main}

\begin{thebibliography}{22}
\providecommand{\natexlab}[1]{#1}
\providecommand{\url}[1]{\texttt{#1}}
\expandafter\ifx\csname urlstyle\endcsname\relax
  \providecommand{\doi}[1]{doi: #1}\else
  \providecommand{\doi}{doi: \begingroup \urlstyle{rm}\Url}\fi

\bibitem[Breiman(1996)]{breiman1996stacked}
Leo Breiman.
\newblock Stacked regressions.
\newblock \emph{Machine learning}, 24\penalty0 (1):\penalty0 49--64, 1996.

\bibitem[D{\v{z}}eroski and {\v{Z}}enko(2004)]{dvzeroski2004combining}
Saso D{\v{z}}eroski and Bernard {\v{Z}}enko.
\newblock Is combining classifiers with stacking better than selecting the best
  one?
\newblock \emph{Machine learning}, 54\penalty0 (3):\penalty0 255--273, 2004.

\bibitem[Dietterich(2000)]{dietterich2000ensemble}
Thomas~G Dietterich.
\newblock Ensemble methods in machine learning.
\newblock In \emph{International workshop on multiple classifier systems},
  pages 1--15. Springer, 2000.

\bibitem[Sill et~al.(2009)Sill, Tak{\'a}cs, Mackey, and Lin]{sill2009feature}
Joseph Sill, G{\'a}bor Tak{\'a}cs, Lester Mackey, and David Lin.
\newblock Feature-weighted linear stacking.
\newblock \emph{arXiv preprint arXiv:0911.0460}, 2009.

\bibitem[Zhou(2012)]{zhou2012ensemble}
Zhi-Hua Zhou.
\newblock \emph{Ensemble methods: foundations and algorithms}.
\newblock CRC press, 2012.

\bibitem[Fan and Gijbels(1992)]{fan1992variable}
Jianqing Fan and Irene Gijbels.
\newblock Variable bandwidth and local linear regression smoothers.
\newblock \emph{The Annals of Statistics}, pages 2008--2036, 1992.

\bibitem[Breiman(2001)]{breiman2001random}
Leo Breiman.
\newblock Random forests.
\newblock \emph{Machine learning}, 45\penalty0 (1):\penalty0 5--32, 2001.

\bibitem[Cs{\'a}ji(2001)]{csaji2001approximation}
Bal{\'a}zs~Csan{\'a}d Cs{\'a}ji.
\newblock Approximation with artificial neural networks.
\newblock \emph{Faculty of Sciences, Etvs Lornd University, Hungary},
  24:\penalty0 48, 2001.

\bibitem[Kingma and Ba(2014)]{adam-optim}
Diederik~P. Kingma and Jimmy Ba.
\newblock Adam: A method for stochastic optimization.
\newblock \emph{CoRR}, abs/1412.6980, 2014.

\bibitem[Glorot and Bengio(2010)]{nn-initialization}
Xavier Glorot and Yoshua Bengio.
\newblock Understanding the difficulty of training deep feedforward neural
  networks.
\newblock \emph{Journal of Machine Learning Research - Proceedings Track},
  9:\penalty0 249--256, 01 2010.

\bibitem[Clevert et~al.(2015)Clevert, Unterthiner, and Hochreiter]{elu}
Djork-Arn\'e Clevert, Thomas Unterthiner, and Sepp Hochreiter.
\newblock Fast and accurate deep network learning by exponential linear units
  (elus), 2015.

\bibitem[Ioffe and Szegedy(2015)]{batch-normalization}
Sergey Ioffe and Christian Szegedy.
\newblock Batch normalization: Accelerating deep network training by reducing
  internal covariate shift.
\newblock In Francis Bach and David Blei, editors, \emph{Proceedings of the
  32nd International Conference on Machine Learning}, volume~37 of
  \emph{Proceedings of Machine Learning Research}, pages 448--456, Lille,
  France, 07 2015. PMLR.
\newblock URL \url{http://proceedings.mlr.press/v37/ioffe15.html}.

\bibitem[Hinton et~al.(2012)Hinton, Srivastava, Krizhevsky, Sutskever, and
  Salakhutdinov]{dropout}
Geoffrey~E. Hinton, Nitish Srivastava, Alex Krizhevsky, Ilya Sutskever, and
  Ruslan~R. Salakhutdinov.
\newblock Improving neural networks by preventing co-adaptation of feature
  detectors.
\newblock \emph{CoRR}, 2012.

\bibitem[Paszke et~al.(2017)Paszke, Gross, Chintala, Chanan, Yang, DeVito, Lin,
  Desmaison, Antiga, and Lerer]{pytorch}
Adam Paszke, Sam Gross, Soumith Chintala, Gregory Chanan, Edward Yang, Zachary
  DeVito, Zeming Lin, Alban Desmaison, Luca Antiga, and Adam Lerer.
\newblock Automatic differentiation in pytorch, 2017.

\bibitem[Nugteren and Codreanu(2015)]{nugteren2015cltune}
Cedric Nugteren and Valeriu Codreanu.
\newblock Cltune: A generic auto-tuner for opencl kernels.
\newblock In \emph{2015 IEEE 9th International Symposium on Embedded
  Multicore/Many-core Systems-on-Chip}, pages 195--202. IEEE, 2015.

\bibitem[Dheeru and Karra~Taniskidou(2017)]{Dua:2017}
Dua Dheeru and Efi Karra~Taniskidou.
\newblock {UCI} machine learning repository, 2017.
\newblock URL \url{http://archive.ics.uci.edu/ml}.

\bibitem[Buza(2014)]{buza2014feedback}
Krisztian Buza.
\newblock Feedback prediction for blogs.
\newblock In \emph{Data analysis, machine learning and knowledge discovery},
  pages 145--152. Springer, 2014.

\bibitem[Hamidieh(2018)]{hamidieh2018data}
Kam Hamidieh.
\newblock A data-driven statistical model for predicting the critical
  temperature of a superconductor.
\newblock \emph{arXiv preprint arXiv:1803.10260}, 2018.

\bibitem[Friedman et~al.(2001)Friedman, Hastie, and
  Tibshirani]{friedman2001elements}
Jerome Friedman, Trevor Hastie, and Robert Tibshirani.
\newblock \emph{The elements of statistical learning}, volume~1.
\newblock Springer series in statistics New York, NY, USA:, 2001.

\bibitem[Meir and R{\"a}tsch(2003)]{meir2003introduction}
Ron Meir and Gunnar R{\"a}tsch.
\newblock An introduction to boosting and leveraging.
\newblock In \emph{Advanced lectures on machine learning}, pages 118--183.
  Springer, 2003.

\bibitem[Pedregosa et~al.(2011)Pedregosa, Varoquaux, Gramfort, Michel, Thirion,
  Grisel, Blondel, Prettenhofer, Weiss, Dubourg, Vanderplas, Passos,
  Cournapeau, Brucher, Perrot, and Duchesnay]{scikit-learn}
F.~Pedregosa, G.~Varoquaux, A.~Gramfort, V.~Michel, B.~Thirion, O.~Grisel,
  M.~Blondel, P.~Prettenhofer, R.~Weiss, V.~Dubourg, J.~Vanderplas, A.~Passos,
  D.~Cournapeau, M.~Brucher, M.~Perrot, and E.~Duchesnay.
\newblock Scikit-learn: Machine learning in {P}ython.
\newblock \emph{Journal of Machine Learning Research}, 12:\penalty0 2825--2830,
  2011.

\bibitem[Van~der Laan et~al.(2007)Van~der Laan, Polley, and
  Hubbard]{van2007super}
Mark~J Van~der Laan, Eric~C Polley, and Alan~E Hubbard.
\newblock Super learner.
\newblock \emph{Statistical applications in genetics and molecular biology},
  6\penalty0 (1), 2007.

\end{thebibliography}

\newpage
\appendix

\section{Proofs}

\textbf{Theorem 3.1.}
Notice that
$$R(G_\theta)=\E \left[ \E\left[\left(Y-G_\theta(\X)\right)^2\right|\X]\right].$$
Hence, in order to minimize $R(G_\theta)$, it suffices to minimize $\E\left[\left(Y-G_\theta(\x)\right)^2\right|\X=\x]$ for
each $\x \in \mathcal{X}$.
Now, once $\sum_{i=1}^k \theta_i(x) = 1$, it follows that,
\begin{align*}
\E\left[\left(Y-G_\theta(\x)\right)^2 |\X=\x\right]&=
\E\left[\left(\sum_{i=1}^k \theta_i(\x)(Y-g_i(\x))  \right)^2  |\X=\x\right]  \\
&=\sum_{i,j} \theta_i(\x)
\theta_j(\x)\E\left[(Y-g_i(\x)) (Y-g_j(\x))  |\X=\x\right]\\
&=
\theta_\x^t \mathbb{M}_\x\theta_\x,
\end{align*}
where $\theta_\x=(\theta_1(\x),\ldots,\theta_k(\x))'$.
Using Lagrange multipliers,
the optimal weights
can by found by minimizing
\begin{align}
\label{eq:f}
f(\theta_\x,\lambda):= 
\theta_\x^t \mathbb{M}_\x\theta_\x - \lambda (\vec{e}^t \theta_\x-1).
\end{align}
Now,
$$\frac{\partial f(\theta_\x,\x)}{\partial \theta_\x} =
2\mathbb{M}_\x\theta_\x -\lambda \vec{e},$$
and therefore the optimal solution
satisfies $\theta^*_\x=\frac{\lambda}{2}\mathbb{M}_\x^{-1} \vec{e}$.
Substituting this on Equation \ref{eq:f} ,
obtain that
$$f(\theta^*_\x,\lambda)=-\frac{\lambda^2}{4}\vec{e}^t 
\mathbb{M}_\x^{-1} \vec{e}+\lambda,$$
and hence
$$\frac{\partial f(\theta^*_\x,\lambda)}{\partial  \lambda}=0 \iff \lambda=\frac{2}{\vec{e}^t \mathbb{M}_\x^{-1}\vec{e}},$$
which yields the optimal solution
$$\theta^*_\x=\frac{\lambda}{2}\mathbb{M}_\x^{-1} \vec{e}=
\frac{2}{\vec{e}^t \mathbb{M}_\x^{-1}\vec{e}}\frac{1}{2}\mathbb{M}_\x^{-1} \vec{e}=\frac{\mathbb{M}^{-1}_\x \vec{e}}{\vec{e}^t \mathbb{M}^{-1}_\x\vec{e}}.$$







\end{document}